\pdfoutput=1

\documentclass[11pt]{article}

\usepackage{EMNLP2022}

\usepackage{times}
\usepackage{latexsym}
\usepackage{graphicx}
\usepackage{tabularx}
\usepackage{multirow}
\usepackage{xspace}
\usepackage{xcolor}
\usepackage{subcaption}

\usepackage[T1]{fontenc}

\usepackage[utf8]{inputenc}

\usepackage{microtype}

\usepackage{inconsolata}

\usepackage{bbding}
\usepackage{pifont}
\usepackage{booktabs}
\usepackage{enumitem}
\usepackage{float}
\usepackage{makecell}
\usepackage[normalem]{ulem}
\usepackage{hyperref}
\hypersetup{colorlinks,urlcolor=blue}

\usepackage{tcolorbox}
\tcbuselibrary{listings}
\newtcblisting{color_box}{
  listing only,
  nobeforeafter,
  after={\xspace},
  hbox,
  tcbox raise base,
  fontupper=\ttfamily,
  colback=lightgray,
  colframe=lightgray,
  size=fbox
  }

\newcommand{\method}{\textsc{DElIteraTeR}\xspace}
\newcommand{\iteraterbase}{\textsc{IteraTeR}\xspace}
\newcommand{\dataset}{\textsc{IteraTeR+}\xspace}

\title{
Improving Iterative Text Revision by Learning Where to Edit\\ from Other Revision Tasks
}

\author{Zae Myung Kim$^{1,3*}$\ , Wanyu Du$^2$\ , Vipul Raheja$^3$\ ,  Dhruv Kumar$^3$\ , Dongyeop Kang$^1$ \\
$^1$University of Minnesota, \space $^2$University of Virginia, \space $^3$Grammarly \\
\texttt{\{kim01756,dongyeop\}@umn.edu} \space
\texttt{wd5jq@virginia.edu}, \space \\
\texttt{\{vipul.raheja,dhruv.kumar\}@grammarly.com}
}

\date{}

\begin{document}

\maketitle

\begingroup\def\thefootnote{*}\footnotetext{The work was done while Zae Myung Kim was interning at Grammarly.}\endgroup

\begin{abstract}
Iterative text revision improves text quality by fixing grammatical errors, rephrasing for better readability or contextual appropriateness, or reorganizing sentence structures throughout a document.
Most recent research has focused on understanding and classifying different types of edits in the iterative revision process from human-written text instead of building accurate and robust systems for iterative text revision.
In this work, we aim to build an end-to-end text revision system that can iteratively generate helpful edits by explicitly detecting \textit{editable spans} (where-to-edit) with their corresponding edit intents and then instructing a revision model to revise the detected edit spans.
Leveraging datasets from other related text editing NLP tasks, combined with the specification of editable spans, leads our system to more accurately model the process of iterative text refinement, as evidenced by empirical results and human evaluations.
Our system significantly outperforms previous baselines on our text revision tasks and other standard text revision tasks, including grammatical error correction, text simplification, sentence fusion, and style transfer.
Through extensive qualitative and quantitative analysis, we make vital connections between edit intentions and writing quality, and better computational modeling of iterative text revisions.
\end{abstract}

\section{Introduction}
\label{sec:intro}

Text revision, naturally, is an iterative process. Writers are required to simultaneously and repeatedly comprehend multiple requirements, such as covering the content, and following linguistic norms and discourse conventions, when producing well-written texts \cite{flower1980dynamics, collins1980framework, vaughan-mcdonald-1986-model}.
Most recent text editing studies have either focused on general-purpose text revision \cite{malmi-etal-2019-encode, mallinson-etal-2020-felix, stahlberg-kumar-2020-seq2edits, li-etal-2022-text}, or targeted monolingual sequence transduction tasks individually, such as grammatical error correction (GEC) \cite{awasthi-etal-2019-parallel, omelianchuk-etal-2020-gector, chen-etal-2020-improving-efficiency}, text simplification \cite{dong-etal-2019-editnts, kumar-etal-2020-iterative, omelianchuk-etal-2021-text, agrawal-etal-2021-non}, and text style transfer \cite{madaan-etal-2020-politeness, malmi-etal-2020-unsupervised, reid-zhong-2021-lewis}, among others.


\begin{table}[t]
  \centering
  \small
  \begin{tabular}{@{}p{0.33\textwidth}@{\hskip 2mm}l@{\hskip 1mm}c@{\hskip 1mm}c@{}}
    \toprule
       & \textbf{Tag} & \textbf{Gran.} & \textbf{Iter.}\\
    \midrule
    \textsc{LaserTagger} \cite{malmi-etal-2019-encode} & O & S & \ding{53}   \\    
    \textsc{FELIX} \cite{mallinson-etal-2020-felix} & O &  S\&P & \ding{53}  \\
    \textsc{Seq2Edits} \cite{stahlberg-kumar-2020-seq2edits}& O & S & \ding{53}  \\
    \textsc{IteraTeR} \cite{du-etal-2022-understanding-iterative} & \ding{53} & S\&P & \ding{51}  \\
    \midrule
    \method (Ours)  & I & S\&P & \ding{51} \\
    \bottomrule
  \end{tabular}
  \caption{\label{tab:datasets}
  Comparison with previous works. Gran. for Granularity: S for sentence-level and P for paragraph-level. Iter. for Iterativeness. Tag for the type of Edit Tagging: O for Edit Operations, I for Edit Intentions.
  \vspace{-5mm}
  }
\end{table}

Despite their progress, these works are quite restricted in their generalizability to practical use cases:
(1) They generally rely on learning edit \textit{operations}, such as \texttt{ADD}, \texttt{KEEP}, \texttt{DELETE}, and \texttt{REPLACE}, which fail to account for many nuanced edit operations such as complex phrasal or sentence rewrites such as word reordering \cite{malmi-etal-2019-encode} or other complex paragraph-level edits. A fundamental limitation of these surface-level edit operations is that they fail to capture the underlying intentions behind the resulting edit operations, and hence, do not learn anything about \textit{why} a part of text was edited in a certain way. For example, a certain span of words may be replaced because it is unclear (\textsc{clarity}) or disfluent (\textsc{fluency}); and depending on this edit \textit{intent} (as opposed to superficial edit \textit{operations}), the revised outcome could be different (Section \ref{sec:intent_control}).
(2) These tagging approaches are inherently limited as they have been developed for \textit{sentence}-level editing \cite{mallinson-etal-2020-felix}.
(3) Since most of the aforementioned studies re-purpose existing sentence-level text editing tasks into monolingual tasks, they are unable to understand or reason about the \textit{iterative} nature of revision, which more closely reflects the human revision process.
Table \ref{tab:datasets} summarizes the comparison with previous works.



In this work, we propose \method: A \textit{\textsc{Del}ineate}-\textit{\textsc{E}dit}-\textit{\textsc{I}terate} approach for the task of Iterative Text \textsc{R}evision \cite{du-etal-2022-understanding-iterative}. Our approach is composed of three stages: (1) \textit{Delineate}: We first detect \textit{editable spans}, the spans of text that require edits, along with their desired edit intentions such as coherence and fluency using a span detection model.
(2) \textit{Edit}: A text revision model then generates the revised text conditioned on the detected \textit{editable spans}.
(3) \textit{Iterate}: The system then continues to iteratively revise the text by going back to Stage 1 (Delineate) until it does not generate further edits or reaches a predefined maximum revision depth.

The main difference of \method from \iteraterbase  \cite{du-etal-2022-understanding-iterative} is that the editable spans are detected first before starting surface-level revisions, making revisions more interpretable and controllable.
Also, each editable span is grounded on corresponding edit intentions, providing more nuanced reasoning behind the edit operations.
We also extend \citet{du-etal-2022-understanding-iterative}'s \iteraterbase dataset (we refer to the augmented dataset as \dataset) by incorporating data from other text editing tasks, leading to significant improvements in performance.\footnote{The datasets, codes, and models can be found at \url{https://github.com/vipulraheja/iterater}.}


Our method shows significant improvements on the Iterative Text Editing task, as well as four well-established monolingual text editing tasks: GEC, sentence fusion, split \& rephrase, text simplification, and formality style transfer.

\section{Related Work}
\vspace{-2mm}
Our work is most closely related to \citet{du-etal-2022-understanding-iterative}, who formally introduced the task of Iterative Text Revision by releasing an annotated dataset of iteratively revised texts, and also used it to provide edit suggestions in a human-in-the-loop iterative editing setting \cite{du-etal-2022-read}. 
In both their works, they computationally model the iterative text revision process, leveraging edit intent information by simply appending it to the input text. However, we improve on their modeling formulation, as evidenced by our experimental results in a significant way: instead of simply appending edit intentions at the beginning of any sentence, we provide more fine-grained edit intention information to our text revision model by first detecting the exact spans which require an edit. Moreover, by incorporating edit-intention-specific knowledge from external task-specific datasets, we are able to push the performance further. 

\section{\method}

\begin{table*}[t]
  \centering
  \small
  \begin{tabular}{@{}l@{\hskip 1mm}|l@{\hskip 1mm}|p{0.32\textwidth}@{\hskip 1mm}|p{0.32\textwidth}@{\hskip 1mm}}
    \toprule
     \textbf{Edit Intention} & \textbf{Dataset} & \textbf{Example Input} & \textbf{Example Output} \\
    \midrule
    \multirow{4}{*}{\textsc{Fluency}} & NUCLE 2014 & Technology based on scientific research requires a wide range of knowledge \textcolor{red}{about} the research. & Technology based on scientific research requires a wide range of knowledge \textcolor{teal}{to conduct} the research. \\
    \cmidrule{2-4}
     & Lang-8 & These days, I write my daily schedule \textcolor{red}{on} a notebook. & These days, I write my daily schedule \textcolor{teal}{in} a notebook. \\ 
    \midrule
    \textsc{Coherence} & DiscoFuse & Their flight is \textcolor{red}{weak. They} run quickly through the tree canopy. & Their flight is \textcolor{teal}{weak, but they} run quickly through the tree canopy. \\
    \midrule
    \multirow{4}{*}{\textsc{Clarity}} & NEWSELA & A storm surge is \textcolor{red}{what forecasters consider} a hurricane's most \textcolor{red}{treacherous} aspect. & A storm surge is \textcolor{teal}{considered} a hurricane's most \textcolor{teal}{dangerous} aspect.  \\
    \cmidrule{2-4}
    & WikiLarge & Wyolica is a \textcolor{red}{two-piece group} from Japan. & Wyolica is a \textcolor{teal}{two person band} from Japan. \\
    \cmidrule{2-4}
    & Split and Rephrase & Aaron Deer \textcolor{red}{plays guitar in} Indie rock \textcolor{red}{style whose origins are coming from the} new wave music. & Aaron Deer \textcolor{teal}{is a an Indie rock guitar player. The stylistic origin of indie rock is} new wave music.  \\
    \midrule
    \textsc{Style} & GYAFC & They \textcolor{red}{wouldnt want u stepping in.} & They \textcolor{teal}{would not desire your interference.}  \\
    \bottomrule
  \end{tabular}
  \caption{\label{tab:external-data-instances}
  Examples of data instances from external corpora used to create the augmented \dataset dataset.
  \vspace{-3mm}
  }
\end{table*}

We follow the Iterative Text Revision task as introduced by \citet{du-etal-2022-understanding-iterative}: given a source document $\mathcal{D}^{t-1}$, at each revision depth $t$, a text revision system will apply a set of edits to get the revised document $\mathcal{D}^{t}$.
The system will continue iterating revision until the revised document $\mathcal{D}^{t}$ satisfies a set of predefined stopping criteria, such as reaching a predefined maximum revision depth $t_{max}$, or making no edits between $\mathcal{D}^{t-1}$ and $\mathcal{D}^{t}$.

In this section, we describe \dataset, the augmented version of the iterative text revision dataset \cite{du-etal-2022-understanding-iterative} and the system pipeline for \method.

\subsection{\dataset: Augmented Dataset}
We use the Iterative Text Revision dataset (\iteraterbase) released by \citet{du-etal-2022-read, du-etal-2022-understanding-iterative} as our primary dataset.
Under their dataset taxonomy, each text edit is broadly categorized into one of two groups: \textsc{meaning-changed} and \textsc{non-meaning-changed}.
Further, edits that belong to the latter group are further assigned to one of the following five sub-groups: \textsc{fluency}, \textsc{coherence}, \textsc{clarity}, \textsc{style}, and \textsc{other}.
This taxonomy of \textit{edit intents} reflects writers' general ``intention" when revising formal documents. It allows us to model the purpose behind each edit of texts, providing more in-depth information than just superficial \textit{edit actions} such as \texttt{ADD}, \texttt{KEEP}, and \texttt{DELETE}.


In this work, we build upon the \iteraterbase dataset by gathering data from other similar text editing tasks according to the aforementioned taxonomy as \iteraterbase (Section \ref{sec:data_aug}).

\subsubsection{Pre-processing}\label{sec:preprocessing}
We observe that many edits from the original dataset were actually \textsc{meaning-changed} edits, i.e., the revised text embodied significantly different content from the old text.
This often occurred when the revisions were made at a document level, reorganizing the paragraphs while adding new contents.
In addition, there were also many cases where significant amounts of texts were either added or deleted by the revisions.
Since new content generation is not the scope of our task, we filtered these type of edits by comparing \textit{length ratio} and \textit{character-level similarity} between original and revised strings, discarding nearly 40\% of the dataset.

\subsubsection{Data Augmentation}\label{sec:data_aug}
The \iteraterbase dataset taxonomy is general enough to encompass other text editing tasks.
For example, the datasets from the GEC task can be viewed as datasets for \textsc{fluency}, text simplification task as \textsc{clarity} edits, sentence fusion or splitting as \textsc{coherence}, and formality style transfer as \textsc{style}.
Using this insight, we adopted the following external datasets for our system. These datasets underwent an identical pre-processing routine as the main dataset.

\paragraph{Fluency} We use two prominent corpora for GEC: the NUS Corpus of Learner English (NUCLE) \cite{dahlmeier-etal-2013-building}, which consists of 1,414 essays written by students at the National University of Singapore (NUS); and the
NAIST Lang-8 Corpus of Learner English \cite{tajiri-etal-2012-tense}, which is one of the largest and most widely used datasets for GEC.
The essays in the NUCLE Corpus were responses to some prompts from various topics including technology innovation and health care, and were hand-corrected by professional English instructors. Lang-8 Corpus, on the other hand, was created by language learners correcting each other's texts. Although these datasets contain multiple fine-grained error categories specific to GEC, in this work, we consider all errors in these corpora as \textsc{fluency}, following the comprehensive definition of fluency in the \iteraterbase dataset taxonomy.

\paragraph{Clarity} We use the Newsela corpus \cite{xu-etal-2015-problems} for the Text Simplification task, which consists of 1,130 articles and their simplified versions which were created by professional editors at Newsela, an online education platform.
We also use WikiLarge, another benchmark dataset for the text simplification task. It was constructed from automatically-aligned complex-simple sentence pairs from English Wikipedia and Simple English Wikipedia \cite{zhu-etal-2010-monolingual, woodsend-lapata-2011-learning, kauchak-2013-improving}. We use the standardized split of this dataset released by \citet{zhang-lapata-2017-sentence} consisting of 296$k$ complex-simple sentence pairs.
Finally, we use the Split and Rephrase \cite{narayan-etal-2017-split} dataset, which includes 1.06M instances mapping a single complex sentence to a sequence of sentences that express the same meaning.
We labeled the edits collected from these datasets as \textsc{clarity} edits.

\paragraph{Coherence} We use the DiscoFuse dataset \cite{geva-etal-2019-discofuse}, which provides a large collection of pairs of sentences that were originally from one coherent sentence, and segmented into two by a rule-based method from sports articles and Wikipedia.
The task then involves linking these two sentences as coherently as possible where it could be done through inserting a discourse connective or merging the input sentences, etc.
These dataset samples were labeled as \textsc{coherence} for our work.

\begin{figure*}[!htbp]
    \includegraphics[width=\textwidth]{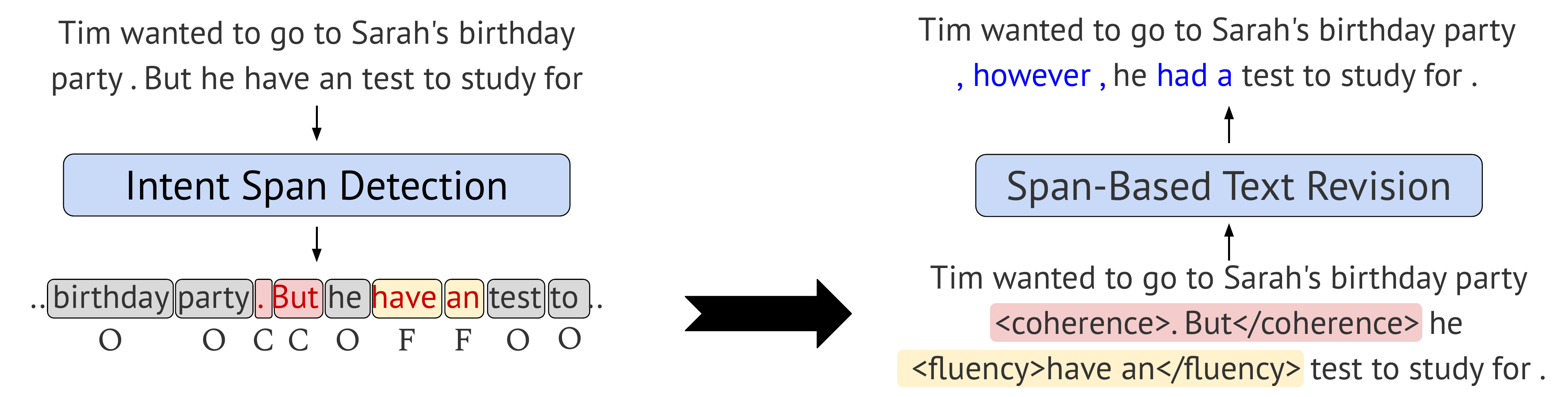}
    \caption{Illustration of the \method pipeline at a given revision depth. While the figure depicts two sentences, the pipeline works for entire paragraphs.}
    \label{fig:pipeline}
\end{figure*}

\paragraph{Style} We use Grammarly’s Yahoo Answers Formality Corpus (GYAFC) \cite{rao-tetreault-2018-dear} which contains 110$k$ informal and formal sentence pairs.
We note that the notion of \textsc{style} edits can be quite subjective as it is about conveying writers' writing preferences, including emotions, tone, and voice; and informal to formal conversion can be viewed as one aspect of \textsc{style}.
We use the dataset for learning informal to formal rewriting because \iteraterbase dataset has primarily been developed for mostly formal writing domains such as ArXiv, Wikipedia and News.

\begin{table}[t!]
\begin{center}
\small
\begin{tabular}{*{4}{l}}
\toprule
\textbf{Intentions} & \textbf{Dataset}  & \textbf{Sentences} & \textbf{Edits} \\
\toprule
\textsc{fluency} &
         \textsc{\iteraterbase} &  131\textit{k} & 131\textit{k} \\
    &      \textsc{task-specific} &  124\textit{k} & 162\textit{k} \\
    \cmidrule{2-4}
    \textsc{clarity}
       &   \textsc{\iteraterbase} &  109\textit{k} & 109\textit{k}  \\
    &      \textsc{task-specific} & 22\textit{k} & 28\textit{k}  \\
    \cmidrule{2-4}
    \textsc{coherence} 
         &  \textsc{\iteraterbase} & 28\textit{k} & 26\textit{k} \\
    &      \textsc{task-specific} & 133\textit{k} & 145\textit{k} \\
    \cmidrule{2-4}
    \textsc{style}
         &  \textsc{\iteraterbase}  & 3\textit{k} & 3\textit{k} \\
    &      \textsc{task-specific} & 45\textit{k} & 90\textit{k} \\
\bottomrule
\end{tabular}
\caption{Data statistics of \dataset. Table \ref{tab:data-stats_full} contains full data statistics.}
\label{tab:data-stats}
\end{center}
\end{table}

Table \ref{tab:data-stats} shows the statistics and intent distributions of all datasets after the pre-processing routine. Table \ref{tab:external-data-instances} depicts instances of data points from all of the external corpora mentioned in this section.

\subsection{System Pipeline}
Our system is arranged in a pipeline where edit intent classification is conducted at token-level as a structured prediction task, followed by span-based text revision where the predicted intent labels are inserted as tag spans in the input.
Figure \ref{fig:pipeline} highlights the overall process of the pipeline at a given revision depth, with an illustrative example. In the iterative text revision setting, this illustrated process repeats until either no editable spans are detected, or a predefined maximum revision depth is reached.

\subsubsection{Intent Span Detection}
Rather than predicting a single edit intent at input-level as done in \citet{du-etal-2022-read}, our model predicts intents at token-level.
Training of such a model was difficult without the construction of our new dataset, since the original \iteraterbase dataset contained a lot of noisy revisions that caused the token-level model to be degenerate.

The intent classification model was trained by fine-tuning a token-level classification layer on top of the pre-trained \textsc{RoBERTa-large} model \cite{liu2019roberta}, where the input to the model is a plain text and output is one of the five classes (\textsc{clarity}, \textsc{coherence}, \textsc{fluency}, \textsc{style}, \textsc{none}) for every token in the input.
We also experimented with a multi-task learning by adding an input-level binary classification layer that predicts if the entire input needs revisions or not.

\begin{table*}[t!]
\begin{center}
\small
\resizebox{\textwidth}{!}{
\begin{tabular}{ll||cccc||cccc||c}
\toprule
& \multirow{2}{*}{\textbf{Training Dataset}} & \multicolumn{4}{c}{\textbf{External Tasks (\method-*)}} & \multicolumn{5}{c}{\textbf{\iteraterbase-\textsc{test}}} \\
\cmidrule{3-11}
& & Clarity & Coherence & Fluency & Style & Clarity & Coherence & Fluency & Style & Overall \\
\toprule
\textsc{SS} & \dataset & \textbf{49.87} & \textbf{98.06} & 78.27 & \textbf{71.89} & \textbf{34.08} & \textbf{24.44} & 63.57 & \textbf{0.42} & 45.99 \\
\textsc{SM} & \dataset & 34.23 & 95.40 & 77.72 & 21.81 & 14.39 & 17.08 & 58.47 & 0.00 & 32.27 \\
\midrule
\textsc{MS} & \dataset & 43.09 & 97.90 & \textbf{79.27} & 65.60 & 33.80 & 22.43 & \textbf{67.36} & 0.00 & \textbf{49.13} \\
\textsc{MM} & \dataset & 43.34 & 96.80 & 77.37 & 71.29 & 33.98 & 20.53 & 64.84 & 0.00 & 47.26 \\
\midrule
\bottomrule
\end{tabular}
}
\caption{Performance of models on intent span detection. All the models are named using the \textsc{XY} convention, where \textsc{X} refers to the Single-sentence (\textsc{S}) vs. Multi-sentence (\textsc{M}) setting and \textsc{Y} refers to the Single-task (\textsc{S}) vs. Multi-task (\textsc{M}) training setting.}
\label{tab:intent_clf_res_token}
\end{center}
\end{table*}
\begin{table*}[t!]
\begin{center}
\small
\resizebox{\textwidth}{!}{
\begin{tabular}{@{}l @{\hspace{0.05cm}}
c@{\hspace{0.1cm}}c@{\hspace{0.1cm}}c
@{\hspace{0.3cm}}
c@{\hspace{0.1cm}}c@{\hspace{0.1cm}}c
@{\hspace{0.3cm}}
c@{\hspace{0.1cm}}c@{\hspace{0.1cm}}c
@{\hspace{0.3cm}}
c@{\hspace{0.1cm}}c@{\hspace{0.1cm}}c
|
c@{\hspace{0.1cm}}c@{\hspace{0.1cm}}c@{}}
\toprule
& \multicolumn{3}{@{}c@{}}{\textbf{\textsc{Clarity-test}}} & \multicolumn{3}{@{}c@{}}{\textbf{\textsc{Coherence-test}}} & \multicolumn{3}{@{}c@{}}{\textbf{\textsc{Fluency-test}}} & \multicolumn{3}{@{}c@{}}{\textbf{\textsc{Style-test}}} & \multicolumn{3}{@{}c@{}}{\textbf{\textsc{\iteraterbase-test}}} \\
\toprule
 & B & R & S & B & R & S & B & R & S & B & R & S & B & R & S \\
\toprule
No Edits Baseline & 0.59 & 74.44 & 23.60 & 0.84 & 97.13 & 31.30 & 0.75 & 88.43 & 25.96 & 0.28 & 61.26 & 15.51 & 0.86 & 91.80 & 29.88 \\
\midrule
\textsc{\iteraterbase-single} & 0.59 & 74.29 & 27.50 & 0.71 & 89.14 & 33.79 & 0.76 & 88.23 & 36.39 & 0.29 & 61.34 & 18.90 & 0.84 & 91.96 & 35.62 \\
\textsc{\iteraterbase-multi} & 0.59 & 74.29 & 27.50 & 0.71 & 89.14 & 33.79 & 0.76 & 88.23 & 36.39 & 0.29 & 61.34 & 18.90 & 0.87 & 93.19 & 43.22 \\
\midrule
\midrule
\textsc{\method-clarity} & 0.62 & 75.93 & 36.63 & 0.34 & 62.79 & 22.60 & 0.25 & 52.14 & 27.06 & 0.04 & 23.01 & 28.28 & 0.55 & 72.14 & 26.43 \\
\textsc{\method-coherence} & 0.17 & 40.52 & 26.33 & 0.96 & 98.74 & 81.17 & 0.24 & 52.20 & 26.72 & 0.03 & 23.79 & 29.81 & 0.39 & 59.74 & 22.60 \\
\textsc{\method-fluency} & 0.16 & 40.29 & 25.93 & 0.52 & 77.23 & 29.09 & 0.86 & 92.38 & 70.83 & 0.04 & 25.08 & 27.66 & 0.61 & 75.15 & 35.18 \\
\textsc{\method-style} & 0.33 & 59.82 & 29.93 & 0.40 & 72.31 & 26.82 & 0.42 & 73.63 & 35.31 & 0.42 & 68.35 & 51.60 & 0.35 & 65.34 & 24.53 \\
\textsc{\method-\iteraterbase} & 0.60 & 75.95 & 51.48 & 0.85 & 95.58 & 51.08 & 0.83 & 90.44 & 61.49 & 0.28 & 56.36 & 36.99 & \textbf{0.92} & \textbf{96.18} & 62.54 \\
\midrule
\textsc{DelIteraTeR-single} & 0.65 & 79.05 & 57.48 & 0.96 & 98.66 & 80.81 & \textbf{0.87} & 92.98 & 73.06 & 0.48 & 71.72 & 60.45 & \textbf{0.92} & 96.14 & 62.06 \\
\textsc{DelIteraTeR-multi} & \textbf{0.66} & \textbf{79.36} & \textbf{58.70} & \textbf{0.96} & \textbf{98.73} & \textbf{81.23} & \textbf{0.87} & \textbf{93.10} & \textbf{73.95} & \textbf{0.49} & \textbf{72.13} & \textbf{61.44} & \textbf{0.92} & 96.13 & \textbf{64.09} \\
\bottomrule
\end{tabular}
}
\caption{Comparison of end-to-end Iterative Text Revision models. B is BLEU, R is ROUGE-L, and S is SARI.}
\label{tab:quant-results}
\end{center}
\end{table*}

\subsubsection{Span-Based Text Revision}
\label{sec:span_text_revision}
The token-level predictions of the intent span prediction model are turned into intent-annotated spans where the part of text that is predicted to be edited is surrounded with intent tags as shown in Figure \ref{fig:pipeline}.
This way, the revision model can focus on parts of the input that need revisions. Note that it is possible to have multiple intent spans in which case the model revises multiple parts of the input.

Following \citet{du-etal-2022-understanding-iterative}, we also fine-tune a \textsc{Pegasus} model \cite{zhang2020pegasus} which is a Transformer-based \cite{vaswani-2017-attention} sequence-to-sequence (Seq2Seq) model.
While more lightweight non-autoregressive models such as \textsc{FELIX} \cite{mallinson-etal-2020-felix} are available, we opted for using the autoregressive Seq2Seq models as our main choice of models.
This is because the task of generating text from editable spans is more involved than generating text from edit operations, showing better performance as described in \citet{du-etal-2022-understanding-iterative}.
\section{Quantitative Results}
While the recent previous studies on text revision include FELIX \cite{mallinson-etal-2020-felix}, LaserTagger \cite{malmi-etal-2019-encode}, Seq2Edits \cite{stahlberg-kumar-2020-seq2edits}, and \textsc{\iteraterbase} \citet{du-etal-2022-understanding-iterative}, we mainly compare our system against the latest work, \textsc{\iteraterbase}, as it had shown consistent improvements over the other aforementioned systems in revision quality \cite{du-etal-2022-understanding-iterative}.

We evaluate our system, \textsc{\method}, on the test splits of \textsc{\dataset} dataset which consists of the original \textsc{\iteraterbase} set and newly augmented task-specific datasets from four different text editing NLP tasks: text simplification, sentence fusion \& splitting, GEC, and formality style transfer, as described in Section \ref{sec:data_aug}.

\subsection{Intent Span Detection}\label{sec:res_intent_clf}

We hypothesize that the prediction of edit intention for a span may benefit from the use of information needed to predict whether an edit is needed or not. For instance, the prediction of a fluency edit may benefit from the detection of a grammatical error in the text. With this idea, we train a \textsc{RoBERTa-Large} model with two different settings:
\begin{enumerate}[noitemsep,topsep=0pt,parsep=0pt,partopsep=0pt]
    \item \textit{Single-Task}: trained for token-level edit intention classification.
    \item \textit{Multi-Task}: Different task-specific heads trained for each task (binary classification task of edit detection, and multi-class edit intention classification).
\end{enumerate}

We also experimented with varying lengths of context windows for the edit intent prediction:
\begin{enumerate}[noitemsep,topsep=0pt,parsep=0pt,partopsep=0pt]
    \item \textit{Single-Sentence}: Only the tokens belonging to a sentence are classified at a given time.
    \item \textit{Multi-Sentence}: For a given sentence for which the intent span detection needs to happen, we concatenate the preceding and succeeding sentence before and after it respectively, to provide additional context to the classification model.
\end{enumerate}

Table \ref{tab:intent_clf_res_token} shows a breakdown of the models' performance on all combinations of the single/multi-sentence and single/multi-task settings for the  intent span detection task. We report F1 scores on both the \iteraterbase dataset, and the test splits of \textsc{task-specific} external datasets which we incorporated into \dataset. We find that the model with multi-sentence, single-task (\textsc{MS}) setting is the best performing one overall. We use this model as our main intent span detection model.
In general, we find that multi-sentence models (\textsc{MM, MS}) perform better than single-sentence models (\textsc{SS, SM}). This can be attributed to the fact that multi-sentence models have access to more context in their inputs, and are able to leverage that to predict edits more accurately. We do notice that the single-sentence single-task (\textsc{SS}) model performs better on the task-specific test sets, and that is attributed to the fact that these datasets are all sentence-level, and did not contain multiple sentences for any added context.

\subsection{Span-Based Text Revision}
As mentioned in Section \ref{sec:span_text_revision}, our main model (\method) for span-based text revision was trained from \textsc{Pegasus-Large} model, following \citet{du-etal-2022-understanding-iterative}.
While we do not bring in additional modeling components to the \textsc{Pegasus} model, we emphasize that the successful training of the model and its performance were heavily dependent on how we constructed the inputs to the model, i.e, sentence-level vs. token-level (span-based) intent annotation.
Specifically, the input to the baseline model, \iteraterbase, was prepared by adding a sentence-level intent class at the beginning of the sentence, whereas \method takes inputs with intent information annotated as tags within corresponding parts of the inputs as shown in Fig.~\ref{fig:pipeline}.
In addition, similar to the intent span detection (Section  \ref{sec:res_intent_clf}), we experimented with both single-sentence (\textsc{single}) and multi-sentence (\textsc{multi}) settings.

In Table \ref{tab:quant-results}, we report the performance of each model using three automatic metrics: BLEU \cite{papineni-etal-2002-bleu}, ROUGE-L \cite{lin-2004-rouge}, and SARI \cite{Xu-etal-sari}.
We observe that span-based \method models outperform the sentence-level \iteraterbase models on all test sets.
We also see that the \textsc{*-multi} models are generally better than \textsc{*-single} models, and this difference is more prominent in \iteraterbase-test\footnote{We note that all of the test sets that we used are filtered as specified in Sec. \ref{sec:preprocessing}} as the test set can cater for multi-sentence inputs while other task-specific sets do not.

We also note that the performance of models was greatly influenced by the \dataset dataset that we collected and pre-processed.
To this end, we report the results of an ablation study on different portions of the dataset in the mid-section of Table \ref{tab:quant-results}, where  \textsc{\method-\{clarity, coherence, fluency, style\}} models were trained on each task-specific dataset only, while \textsc{\method-\iteraterbase} model was trained on the filtered version of \iteraterbase dataset.
The results show that while the individual task-specific models perform well on their corresponding test sets, the quality of revision drops significantly when tested on the \iteraterbase-test.
Similarly, the model that was solely trained on \iteraterbase dataset does not perform well on the task-specific test sets.
As expected, the best performance was achieved when the models were trained on the full \dataset dataset.

\section{Qualitative Results}
In this section, we analyze the behavior of our \method system in greater detail by looking at the development of edit intent trajectories as we run the system iteratively using its previous revision $\mathcal{D}^{t-1}$ to generate the revision at depth $t$ (Section \ref{sec:intent_trajec}).
In Section \ref{sec:intent_control}, we try to probe our text revision model by modifying the outputs of the span-based model to see if we can generate diverse revisions by placing (1) different intents on the same edit spans, and (2) same intent on different edit spans.

\subsection{Edit intention trajectories by depths}\label{sec:intent_trajec}
To understand if there is a clear pattern in trajectories of edit intents with the progression of revision, Sankey diagrams were drawn for revision results ($t<5$) on each test set.
At revision depth $t$, we record the number of instances of all possible intent transitions, $\textsc{intent}_{i}^{t-1} \rightarrow \textsc{intent}_{j}^{t}$, defining the flow for the Sankey diagram.

\begin{figure}[!tbp]
\vspace{-5mm}
    \includegraphics[width=\columnwidth]{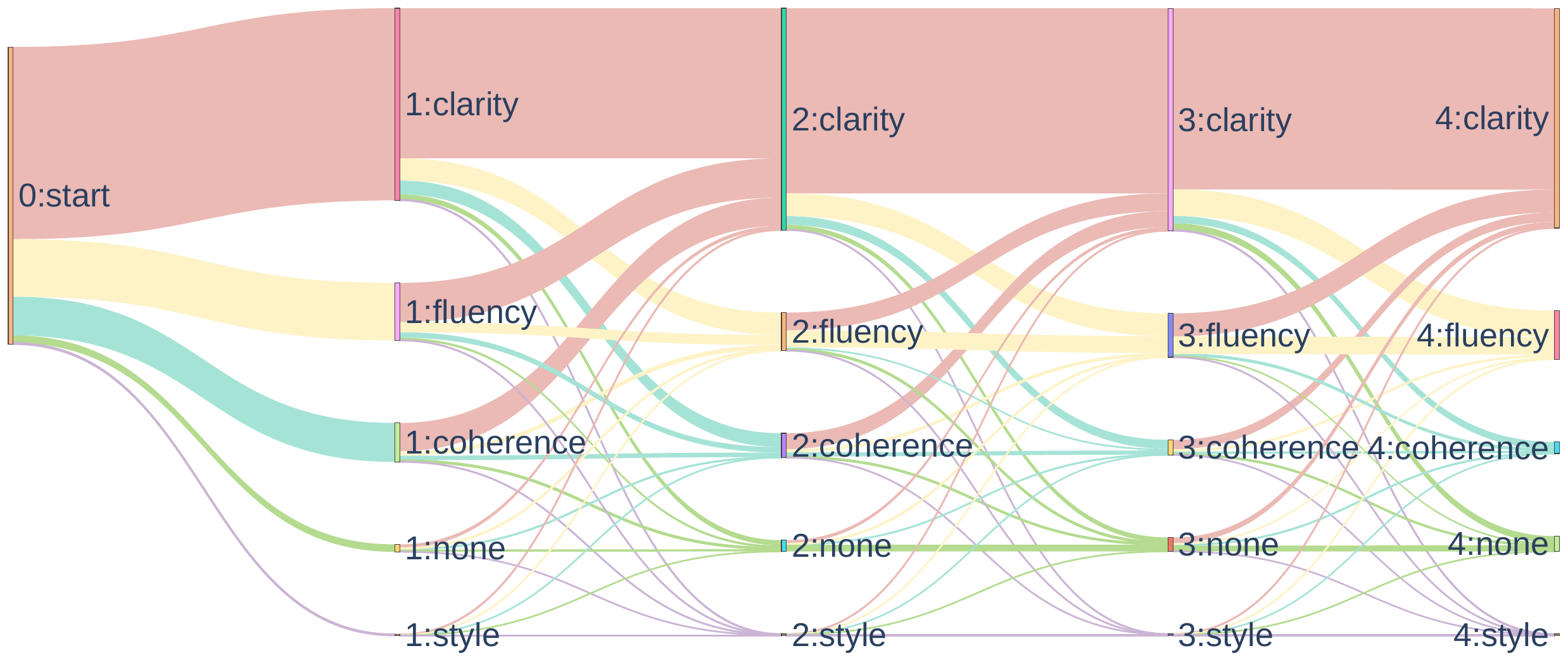}
    \vspace{-5mm}
    \caption{Illustration of intention trajectories by \method for iterative revision on \textsc{\iteraterbase-test}.}
    \label{fig:sankey_200K}
\end{figure}

\begin{figure*}[!htbp]
    \begin{minipage}[c]{\columnwidth}
        \centering
        \includegraphics[width=\columnwidth]{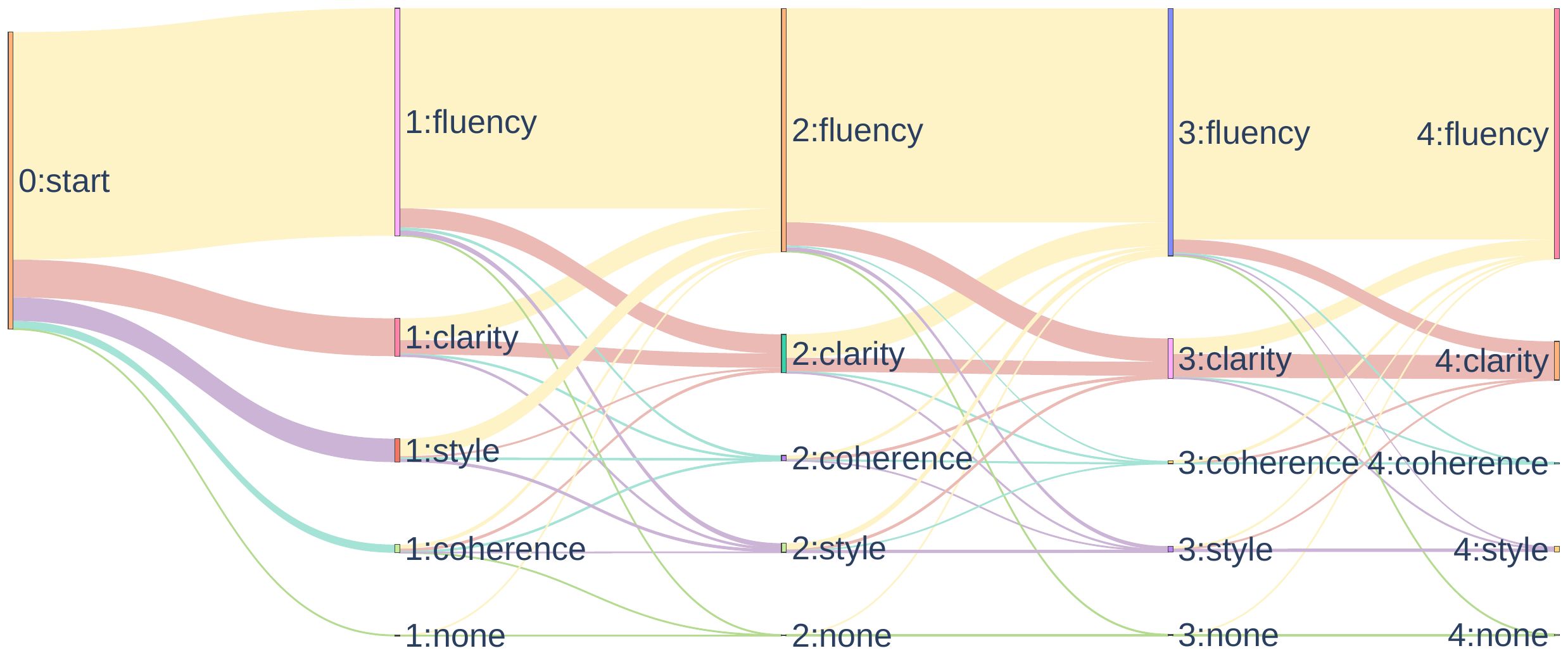}
        \subcaption{Essays written by low level of English skills}
    \end{minipage}\hfill
    \begin{minipage}[c]{\columnwidth}
        \centering
        \includegraphics[width=\columnwidth]{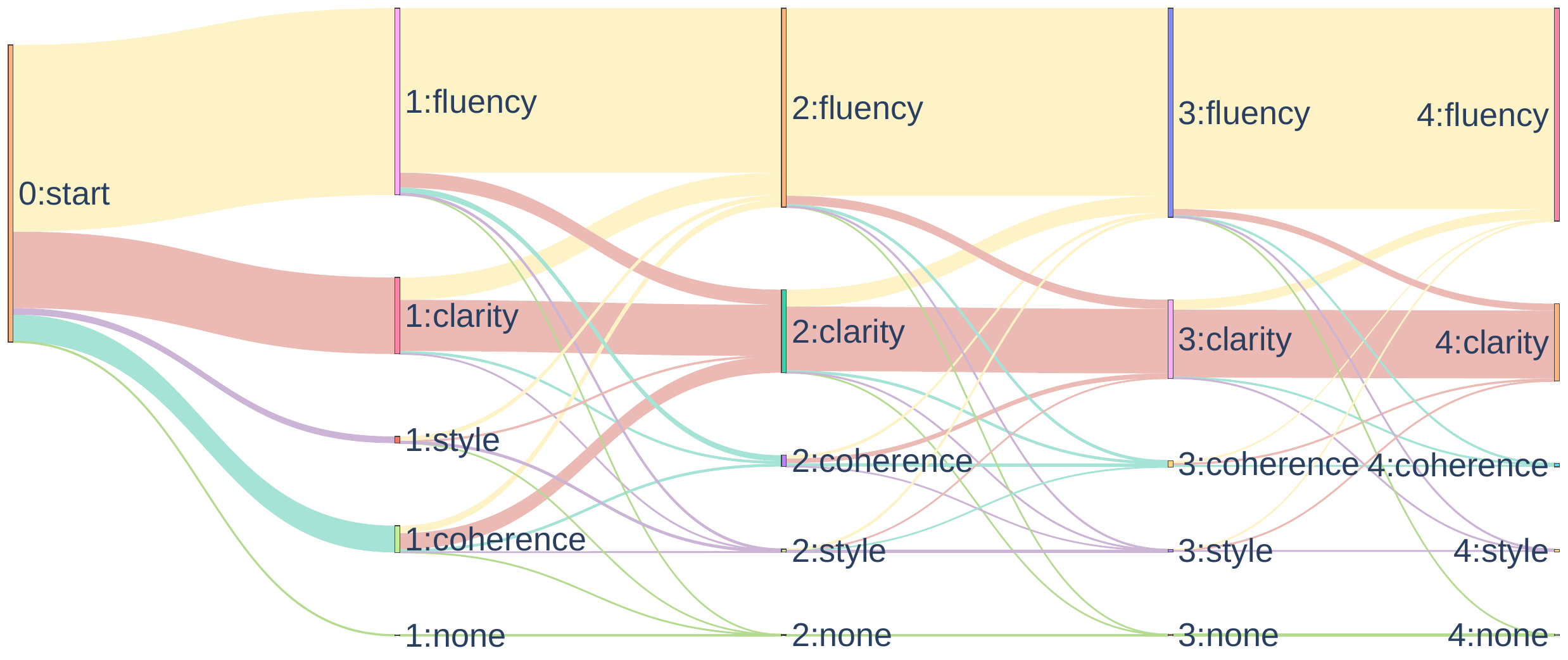}
        \subcaption{Essays written by high level of English skills}
    \end{minipage}
    \caption{Sankey diagrams on ETS Corpus of Non-Native Written English.}
    \label{fig:sankey_ets}
\end{figure*}

\begin{table}[t]
\resizebox{\columnwidth}{!}{
\small
\begin{tabular}{@{}p{0.5\columnwidth} | p{0.45\columnwidth}@{}}
\toprule
\textbf{Revision Input} & \textbf{Revision Output} \\
\midrule

I \textcolor{purple}{\textbf{\texttt{<fluency>}}} disagree about that "young people do not give enough time to helping their communities" \textcolor{purple}{\textbf{\texttt{</fluency>}}}. & 
I disagree with the statement that "young people do not give enough time to helping their communities". \\
\midrule

I \textcolor{purple}{\textbf{\texttt{<clarity>}}} disagree about that "young people do not give enough time to helping their communities" \textcolor{purple}{\textbf{\texttt{</clarity>}}}. & 
I disagree that young people do not give enough time to helping their communities. \\
\midrule
 I \textcolor{purple}{\textbf{\texttt{<coherence>}}} disagree about that "young people do not give enough time to helping their communities" \textcolor{purple}{\textbf{\texttt{</coherence>}}}. & 
I disagree. \\
\midrule
I \textcolor{purple}{\textbf{\texttt{<style>}}} disagree about that "young people do not give enough time to helping their communities" \textcolor{purple}{\textbf{\texttt{</style>}}}. & 
I disagree with the statement that "young people do not give enough time to helping their communities". \\
\bottomrule
\end{tabular}
}
\caption{\label{tab:samespan_diffedits}
Effect of detected edit intentions on generated revisions.}
\vspace{-0.5cm}
\end{table}
\begin{table}[t]
\resizebox{\columnwidth}{!}{
\small
\begin{tabular}{@{}p{0.52\columnwidth} | p{0.45\columnwidth}@{}}
\toprule
\textbf{Revision Input} & \textbf{Revision Output} \\
\midrule
    I \texttt{<fluency>}\textcolor{purple}{disagree about that "young people do not give enough time to helping their communities"}\texttt{</fluency>}. & I \textcolor{teal}{disagree with the statement that "young people do not give enough time to helping their communities"}. \\
\midrule
    I \texttt{<fluency>}\textcolor{purple}{disagree about}\texttt{</fluency>} that "young people do not give enough time to helping their communities". & I \textcolor{teal}{disagree with} that "young people do not give enough time to helping their communities".\\
\midrule
    I disagree about that "young people \texttt{<clarity>}\textcolor{purple}{do not give enough time}\texttt{</clarity>} to helping their communities". 
    & I disagree about that "young people \textcolor{teal}{do not have enough time} to helping their communities". \\
\midrule
    I disagree about that "young people do not give enough time \texttt{<clarity>}\textcolor{purple}{to helping their communities}\texttt{</clarity>}". & I disagree about that "young people do not give enough time \textcolor{teal}{to help their communities}."\\
\bottomrule
\end{tabular}
}
\caption{\label{tab:sameintent_diffspan}
Effect of detected edit spans (for the same intents) on generated revisions.}
\vspace{-0.5cm}
\end{table}

Figure \ref{fig:sankey_200K} shows a Sankey diagram for \iteraterbase-test dataset.\footnote{The rest of the diagrams are added to the Appendix \ref{appendix:edit-intent-trajectories}}
On this test set, we can observe that most of the intent sequences are flowing from \textsc{clarity} to \textsc{clarity}, followed by edit sequences that go into \textsc{fluency}.
The diagram is, of course, influenced by the distribution of edit intents present in the test set.
With the task-specific test sets, we confirm that the corresponding intents tend to be the major flow in the diagrams.
This begs the question: would we observe similar results with documents from the same domain but written by writers with different levels of English proficiency?

To answer this question, we performed a similar experiment using English essays from the ETS Corpus of Non-Native Written English \cite{ets-dataset} where writers with different levels of English proficiency answered prompt questions by composing short essays.
Using the validation and test sets of the corpus, we gathered 261 essays separately from two groups: (a) writers with low English proficiency vs. (b) high English proficiency.
These essays answer 8 different essay topics, and the number of essays for each topic is kept the same between the groups.

Figure \ref{fig:sankey_ets} shows the Sankey diagrams for both groups, computed using their essays and the corresponding iterative revisions.
In the figure, we can identify some distinctive patterns in the flows of edit intents; for example, more \textsc{fluency} edit transitions occur for group (a) than (b) which can be considered as more superficial errors than other intent types.
For group (b), we see more \textsc{coherence} errors that later make transition to \textsc{clarity}.

This finding suggests that there could be an optimal path of edit transition for a given document of a particular domain; which could be learned by a reinforcement learning algorithm where a reward would be a combination of automatic scores of revised documents: something we would like to explore in future work.

So far, at the current state of the models, we do sometimes observe some pitfalls where continuously applying the revision model multiple times either degrades the quality of the text by removing words or gets stuck in a deadlock position where the model oscillates between applying and reverting a single revision. The former can be mitigated by stopping the revision process when a score from an automatic metric such as SARI decreases. The latter deadlock case can be filtered heuristically as well. We point out that these phenomena are also observed in other text revision models, especially more so when their training datasets include a lot of meaning-changed edits that are deletions.

\begin{figure}[!tbp]
    \includegraphics[width=\columnwidth]{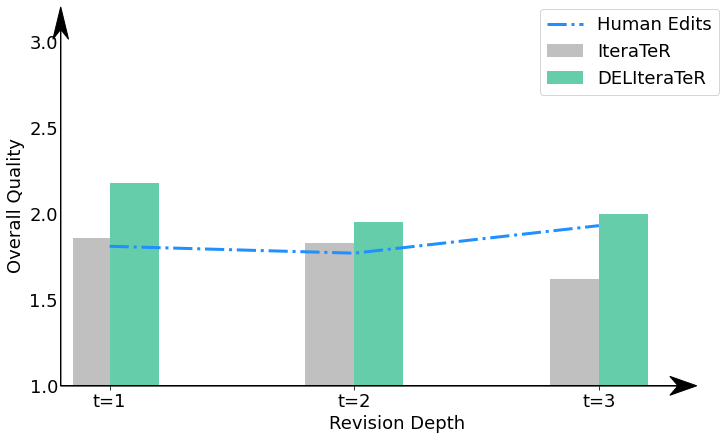}
    \vspace{-5mm}
    \caption{Manual pair-wise comparison for document revisions without \textsc{meaning-changed} edits.}
    \label{fig:human_overall_bar}
\end{figure}

\subsection{Effect of intents and spans on revisions}\label{sec:intent_control}
We also try to probe the system to understand the behaviors of the constituent systems. In particular, the sensitivity of the revision generation model to the outputs of the intent span detection model. To do this, we conducted two analyses: (1) We modified the edit intentions keeping the editable span the same; and (2) while keeping the predicted edit intention the same, we varied the editable span.
By modifying the inputs to the text revision model in this manner, we analyzed its outputs.

In Table \ref{tab:samespan_diffedits}, we can observe that placing a \textsc{fluency} intent span revises the less fluent original sentence by linking a correct preposition for the verb ``disagree'' and adding the following noun, ``statement''. Similarly, a \textsc{clarity} intent simplifies the sentence by merging the quoted segment. However, as shown with the results from \textsc{coherence} and \textsc{style} intents in the table, the revised outputs may not always be preferable.

Table \ref{tab:sameintent_diffspan} shows the effect of varying the length of editable spans. We see that if there is a change in the original sentence (i.e. a revision is predicted), that change only occurs within the bounds of the editable span.
The results obtained with the editable intent spans suggest that we can control and influence the model's generations to a certain extent.
\section{Human Evaluation}
To better understand how our system affects the text quality and the iterative revision process, we conducted human evaluations to investigate how do text editing models affect document quality. 

We hired a group of proficient linguists to evaluate the quality of the documents being edited across multiple (up to 3) revisions, where each revision was annotated by 3 linguists. For each revision, we randomly shuffle the original and revised texts, and ask the evaluators to select which one was better in terms of fluency, coherence, readability, meaning preservation, and overall quality. They could choose one of the two texts, or neither. Then, we calculated the score for the quality of the human revisions as follows: 1 means the revised text is worse compared to the original text; 2 means the revised text does not show a better quality than the original text, or there was no agreement among the 3 annotators; 3 means that the revised text was better than the original text.

Figure \ref{fig:human_overall_bar} shows the results of the human evaluation on the aforementioned criteria. We choose our best-performing model (\method-\textsc{multi}) trained on \dataset using the \textit{delineate-edit-iterate} approach to generate revisions by first identifying editable spans, and compare with human revisions and the text revision model from \citet{du-etal-2022-understanding-iterative}. 
We see that our system produces the best overall results, outperforming the human edits, as well as \iteraterbase system in overall quality. This is a major improvement relative to \cite{du-etal-2022-understanding-iterative} where model revisions were significantly underperformed by human edits in overall quality.
Table \ref{tab:revision_example} shows an example of iterative text revision generated by \iteraterbase and \method, respectively.
\section{Conclusion and Discussions}
We propose \textsc{\method}: an improved system for Iterative Text Revision, using a \textit{delineate-edit-iterate} framework, consisting of an intent span detection model, and a text revision generation model, based on the \iteraterbase framework of \citet{du-etal-2022-understanding-iterative}.
The edit intent detection model is a token-level edit-intention classification model which detects \textit{editable spans}: spans of text that require an edit along with the type of edit needed.
The text revision model is a generative model, which makes revisions to the detected editable spans, conditioned on their corresponding edit intentions. 
We also create \dataset: an expanded version of \cite{du-etal-2022-understanding-iterative} \iteraterbase dataset by incorporating data from other text editing NLP tasks. 
Leveraging this dataset and the \textit{delineate-edit-iterate} framework, our system supervises the revision generation model to reflect both the location, and intentions behind the desired revision, leading to superior performance on the Iterative Text Revision task, compared to other baselines and related works. 

Experiments on the standard \iteraterbase test dataset, as well as standard NLP text editing datasets demonstrate the effectiveness of our framework for the task. Moreover, human evaluations indicate that our system produces the best overall results, outperforming the human edits, as well as \iteraterbase system across all revision depths.
Additionally, we provide insights into our models by probing them, and into the progression of iterative text revision by analyzing the edit intent trajectories across both our test datasets as well as a dataset of English essays from the ETS corpus, hinting the possibility of learning optimal revision paths possibly thorough reinforcement learning. In the future, we plan to investigate in this direction as well as improving the general robustness of the system by task-specific data augmentation with induced noise.
\section{Limitations}
We note that the augmented task-specific datasets were only available at sentence-level.
While the augmentation did improve the models' performance on \iteraterbase-test set, it still lacked the contextual information to to unlock the full potential of multi-sentence modeling.
Also, while we conducted user studies on the quality of the generated revisions, our current version of work does not yet provide results obtained with \textit{human-in-the-loop} deployment where users are involved in the iterative revision process along with the revision system.
Another limitation of our work is that the revision system is geared toward generating formal writing than informal and casual writing.
\section{Ethical Considerations}
All the data collected in this work is from publicly available sources, and the original document authors' copyrights are respected. During the data annotation process, all human evaluators are anonymized to respect their privacy rights. All human evaluators get a fair wage that is higher than the minimum wage based on the number of data points they evaluate. There is no risk that the harms of our work will disproportionately fall on marginalized or vulnerable populations. Our datasets do not contain any identity characteristics (e.g. gender, race, ethnicity), and will not have ethical implications of categorizing people.

In terms of our models, we recognize that by using text generation models as part of our system, they are susceptible to issues of hallucination and other potentially harmful content \cite{maynez-etal-2020-faithfulness, gehman-etal-2020-realtoxicityprompts}. However, since the focus of our system is on text editing, we are able to mitigate some of these issues by carefully curating our datasets. We ignore any data points which lead to meaning-changing edits, thereby reducing the chances of hallucination, or generation of new and potentially harmful content.

\bibliography{anthology,custom}
\bibliographystyle{acl_natbib}

\clearpage

\appendix
\onecolumn
\section{Statistics on \dataset Dataset}
In Table \ref{tab:data-stats_full} we show detailed statistics of our \dataset dataset, showing before and after the filtering process.
\begin{table}[h!]
\begin{center}
\small
\resizebox{0.7\textwidth}{!}{%
\begin{tabular}{*{7}{l}}
\toprule
\multirow{2}{*}{\textbf{Split}} & \multirow{2}{*}{\textbf{Intent}}  & \multirow{2}{*}{\textbf{Source}} & \multicolumn{2}{c}{\textbf{Before}} & \multicolumn{2}{c}{\textbf{After}} \\
\cmidrule{4-7}
& & & \textbf{Sentences} & \textbf{Edits} & \textbf{Sentences} & \textbf{Edits}\\
\toprule
\multirow{8}{*}{Train} 
    & \multirow{2}{*}{\textsc{fluency}}
         &  \textsc{\iteraterbase} & 142\textit{k} & 142\textit{k} & 107\textit{k} & 107\textit{k} \\
    &    &  \textsc{task-specific} & 1.1M & 779\textit{k} & 122\textit{k} & 158\textit{k} \\
    \cmidrule{2-7}
    & \multirow{2}{*}{\textsc{clarity}}
         &  \textsc{\iteraterbase} & 140\textit{k}  & 140\textit{k} & 90\textit{k} & 90\textit{k}  \\
    &    &  \textsc{task-specific} & 1.69M & 6.07M & 22\textit{k} & 28\textit{k}  \\
    \cmidrule{2-7}
    & \multirow{2}{*}{\textsc{coherence}}    
         &  \textsc{\iteraterbase} & 70\textit{k} & 70\textit{k} & 24\textit{k} & 24\textit{k} \\
    &    &  \textsc{task-specific} & 4.49M & 5.1M & 120\textit{k} & 132\textit{k} \\
    \cmidrule{2-7}
    & \multirow{2}{*}{\textsc{style}}
         &  \textsc{\iteraterbase} & 3\textit{k} & 3\textit{k} & 2.5\textit{k} & 2.5\textit{k} \\
    &    &  \textsc{task-specific} & 104\textit{k} &  246\textit{k} & 29\textit{k} & 60\textit{k} \\
\midrule
\multirow{8}{*}{Valid} 
    & \multirow{2}{*}{\textsc{fluency}}
         &  \textsc{\iteraterbase} & 17\textit{k} & 17\textit{k} & 11\textit{k} & 11\textit{k}  \\
    &    &  \textsc{task-specific} & - & - & 1\textit{k} & 2k\\
    \cmidrule{2-7}
    & \multirow{2}{*}{\textsc{clarity}}
         &  \textsc{\iteraterbase} & 15\textit{k} & 15\textit{k}  & 9\textit{k} & 9.5\textit{k}\\
    &    &  \textsc{task-specific} & 42\textit{k} & 139\textit{k}  & 122 & 170 \\
    \cmidrule{2-7}
    & \multirow{2}{*}{\textsc{coherence}}    
         &  \textsc{\iteraterbase} & 8.7\textit{k} & 8.7\textit{k} & 2.5\textit{k} & 2.5\textit{k} \\
    &    &  \textsc{task-specific} & 46\textit{k} & 52\textit{k} & 1\textit{k} & 1.1\textit{k} \\
    \cmidrule{2-7}
    & \multirow{2}{*}{\textsc{style}}
         &  \textsc{\iteraterbase} & 366 & 366 & 265 & 265  \\
    &    &  \textsc{task-specific} & 41\textit{k} & 92\textit{k}  & 11\textit{k} & 20\textit{k}\\
\midrule
\multirow{8}{*}{Test} 
    & \multirow{2}{*}{\textsc{fluency}}
         &  \textsc{\iteraterbase} & 20\textit{k} & 20\textit{k}  & 13\textit{k} & 13\textit{k} \\
    &    &  \textsc{task-specific} & - & - & 1.5\textit{k} & 1.9\textit{k}\\
    \cmidrule{2-7}
    & \multirow{2}{*}{\textsc{clarity}}
         &  \textsc{\iteraterbase} & 17\textit{k} & 17\textit{k} & 10\textit{k} & 10\textit{k} \\
    &    &  \textsc{task-specific} & 45\textit{k} & 149\textit{k}  & 132 & 200 \\
    \cmidrule{2-7}
    & \multirow{2}{*}{\textsc{coherence}}    
         &  \textsc{\iteraterbase} & 10\textit{k} & 10\textit{k} & 2\textit{k} & 2\textit{k} \\
    &    &  \textsc{task-specific} & 44\textit{k} & 50\textit{k} & 12\textit{k} & 13\textit{k} \\
    \cmidrule{2-7}
    & \multirow{2}{*}{\textsc{style}}
         &  \textsc{\iteraterbase} & 447 & 447  & 330 & 330 \\
    &    &  \textsc{task-specific} & 19\textit{k} & 45\textit{k} & 5\textit{k} & 10\textit{k} \\
\bottomrule
\end{tabular}
}
\caption{Dataset splits and sizes. "Before" refers to the raw data statistics before the pre-processing routine (Section \ref{sec:preprocessing}), and "After" refers to the data statistics after the pre-processing was applied.}
\label{tab:data-stats_full}
\end{center}
\end{table}

\section{Training Details}
Throughout our experiments, we mostly adopted codes released by \citet{du-etal-2022-read, du-etal-2022-understanding-iterative}.
We did not conduct any additional hyper-parameter search, but followed the same hyper-parameter settings as \citet{du-etal-2022-read, du-etal-2022-understanding-iterative} when training both intent classification and text revision models.
We plan to make our version of codes and final datasets publicly available upon acceptance.

We used Transformers library \cite{wolf-etal-2020-transformers} from Hugging Face to train and run the models using four NVIDIA V100 GPUs in a distributed data-parallel setting.
The intent classification models (\textsc{RoBERTa-large}) were generally trained to convergence within 10 hours, while the text revision models (\textsc{Pegasus-large}) took up to five days to train for 3 epochs using our full \dataset dataset.
We note that we did not introduce any extra model parameters, and therefore the network sizes are identical to that of the corresponding original models.
The models were saved every 2,000 batch steps and selected based on the validation performance on the \iteraterbase dataset.

\section{Edit Intention Trajectories}
\label{appendix:edit-intent-trajectories}
\begin{figure*}[!t]
    \centering
    \begin{minipage}[c]{0.75\columnwidth}
        \centering
        \includegraphics[width=\columnwidth]{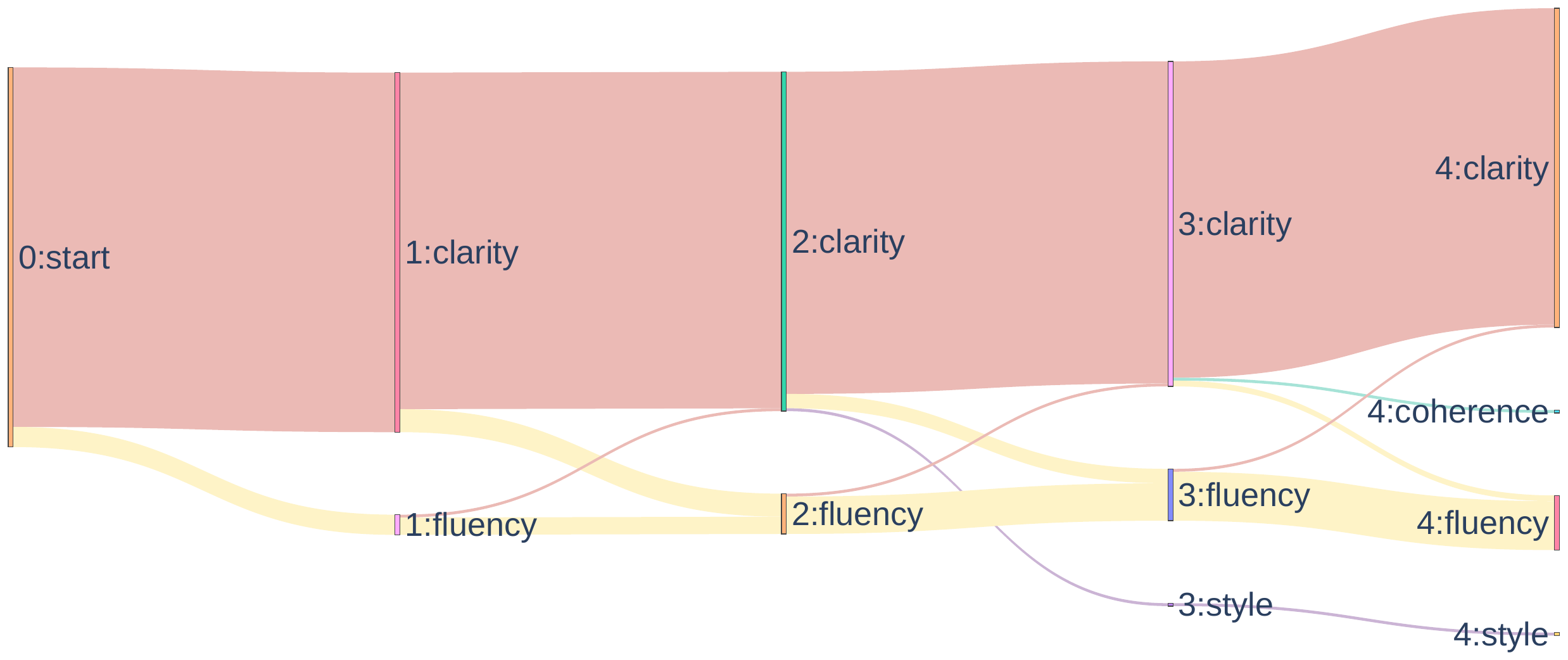}
        \subcaption{\textsc{Clarity}}
    \end{minipage}\vspace{0.5cm}
    \begin{minipage}[c]{0.75\columnwidth}
        \centering
        \includegraphics[width=\columnwidth]{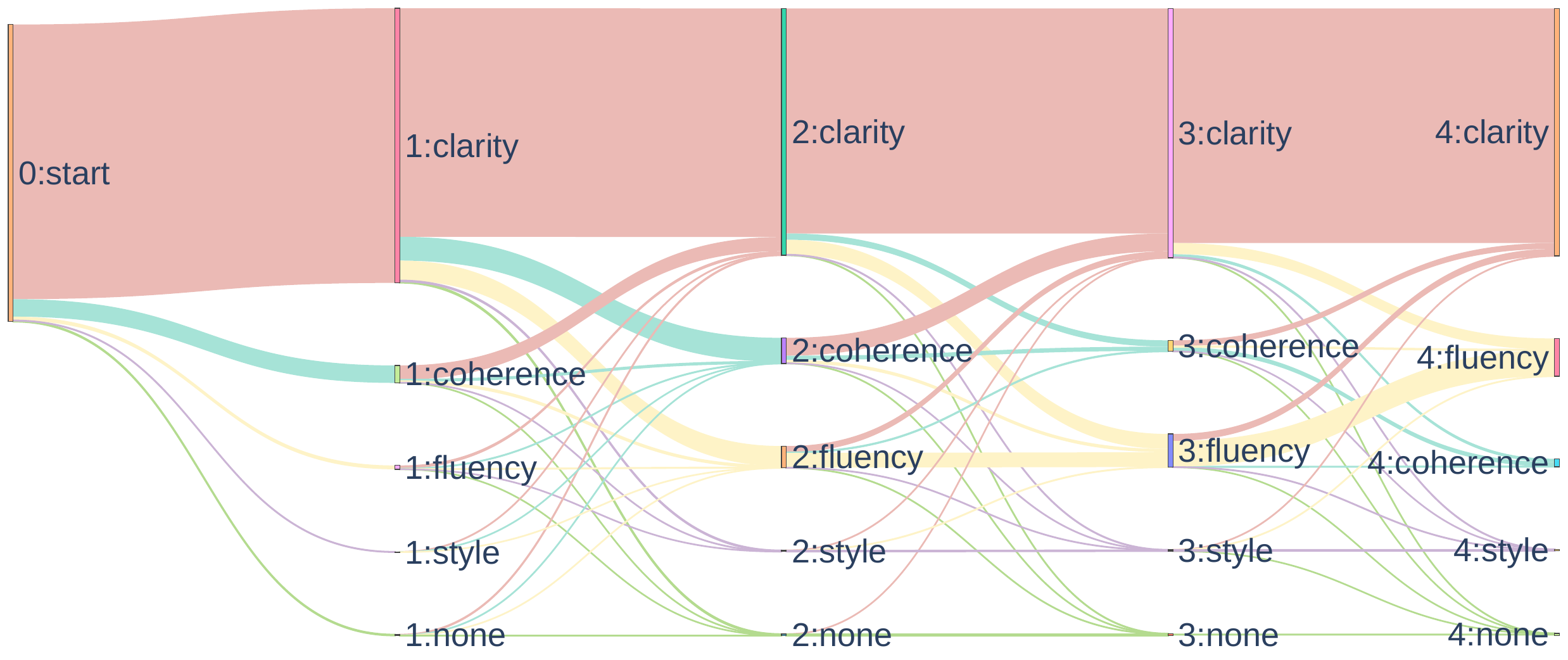}
        \subcaption{\textsc{Coherence}}
    \end{minipage}\vspace{0.5cm}
    \begin{minipage}[c]{0.75\columnwidth}
        \centering
        \includegraphics[width=\columnwidth]{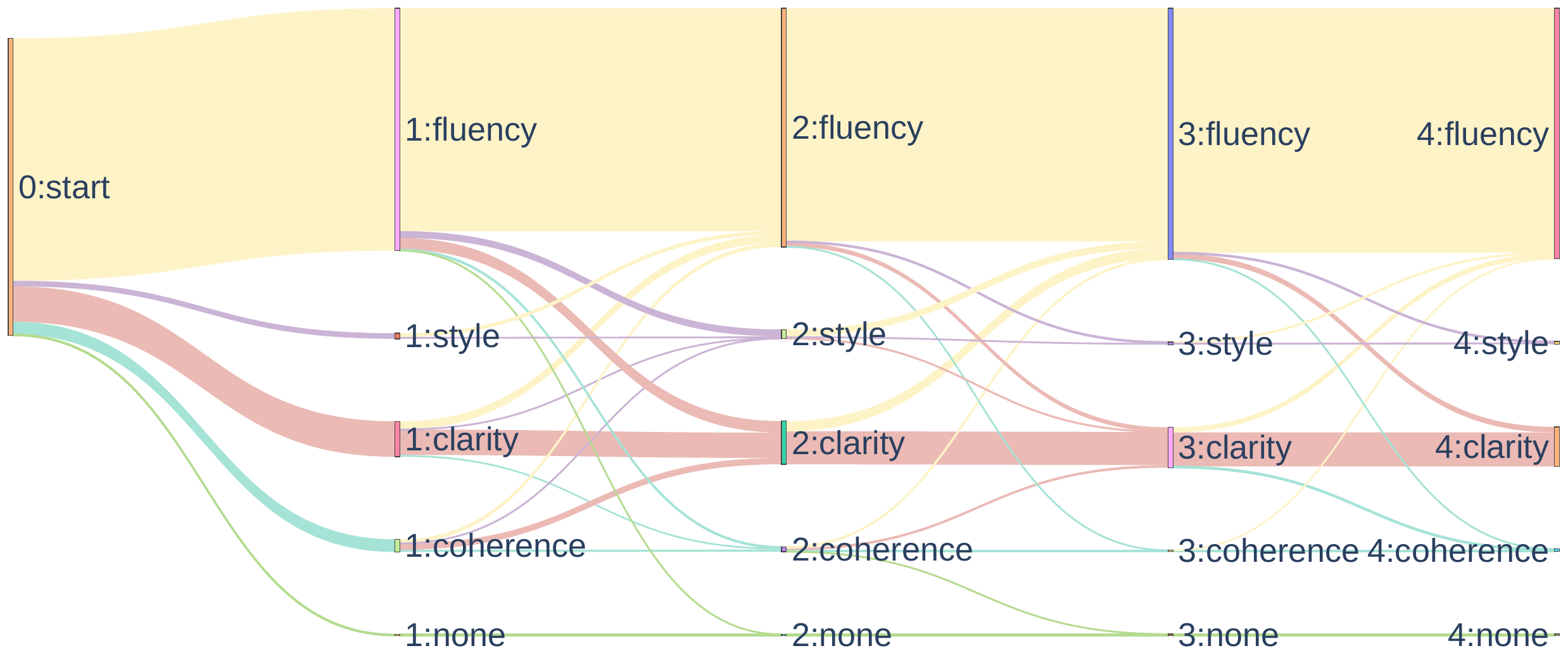}
        \subcaption{\textsc{Fluency}}
    \end{minipage}\vspace{0.5cm}
    \begin{minipage}[c]{0.75\columnwidth}
        \centering
        \includegraphics[width=\columnwidth]{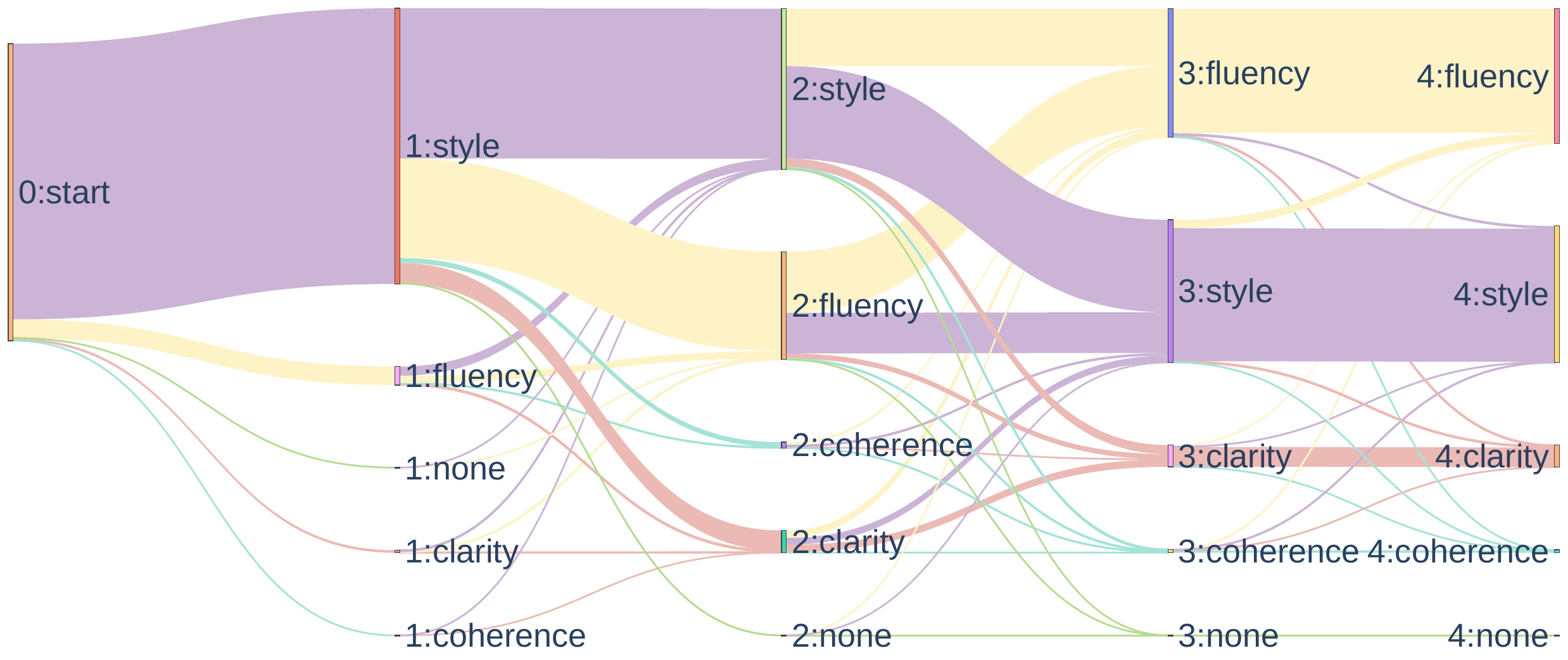}
        \subcaption{\textsc{Style}}
    \end{minipage}
    \caption{Sankey diagrams illustrating edit intention trajectories for task-specific test sets}
    \label{fig:sankey_task_specific}
\end{figure*}

Figure \ref{fig:sankey_task_specific} shows Sankey diagrams drawn for the task-specific test sets of \dataset dataset.

\section{Document-level Revision Examples}
Table \ref{tab:revision_example} shows iterative revisions generated by \iteraterbase and \method (ours), respectively.

\begin{table*}[!t]
  \centering
  \small
  \begin{tabular}{@{}r|p{0.26\textwidth}|p{0.26\textwidth}|p{0.26\textwidth}@{}}
    \toprule
     $t$ & Human Edits & \iteraterbase & \textbf{\method} (ours)  \\
    \midrule
    0 
    & The insurgent maoists would be allowed to return to the government. The treaty establishes that a constitutional asembly should form in April to rewrite the constitution, formally end the monarchy and put together the details of the new republican system. The maoist rebels declared a cease fire and signed a peace treaty, agreeing to place its troops and weapons under U.N. supervision.
    & The insurgent maoists would be allowed to return to the government. The treaty establishes that a constitutional asembly should form in April to rewrite the constitution, formally end the monarchy and put together the details of the new republican system. The maoist rebels declared a cease fire and signed a peace treaty, agreeing to place its troops and weapons under U.N. supervision.
    & The insurgent maoists would be allowed to return to the government. The treaty establishes that a constitutional asembly should form in April to rewrite the constitution, formally end the monarchy and put together the details of the new republican system. The maoist rebels declared a cease fire and signed a peace treaty, agreeing to place its troops and weapons under U.N. supervision. \\
    \midrule
    1 
    & The insurgent maoists would be allowed to return to the government. The treaty establishes that a constitutional asembly should form in April to rewrite the constitution, formally end the monarchy and put together the details of the new republican system. The maoist rebels declared a cease fire and signed a peace treaty, agreeing to place its troops and weapons under \colorbox{red!50}{\sout{U.N.}}\colorbox{teal!50}{United Nations} supervision.
    & The \colorbox{red!50}{\sout{insurgent maoists}}
    \colorbox{teal!50}{insurgents} would be allowed to return to the government. The treaty establishes that a constitutional asembly should \colorbox{red!50}{\sout{form}}\colorbox{teal!50}{be formed} in April to rewrite the constitution, formally end the monarchy and put together the details of the new republican system. The \colorbox{red!50}{\sout{maoist}}\colorbox{teal!50}{Maoist} rebels declared a cease fire and signed a peace treaty, agreeing to place its troops and weapons under U.N. supervision.
    & The \colorbox{red!50}{\sout{insurgent maoists}}
    \colorbox{teal!50}{insurgents} would be allowed to return to the government. \colorbox{red!50}{\sout{The treaty}}\colorbox{teal!50}{It} establishes that a constitutional asembly should form in April to rewrite the constitution, formally end the monarchy and put together the details of the new republican system. \colorbox{red!50}{\sout{The maoist rebels}}\colorbox{teal!50}{Rebels} declared a cease fire and signed a peace treaty, agreeing to place its troops and weapons under U.N. supervision. \\
    \midrule
    2 
    & The insurgent \colorbox{red!50}{\sout{maoists}}\colorbox{teal!50}{Maoists} would be allowed to return to the government. The treaty establishes that a constitutional \colorbox{red!50}{\sout{asembly}}\colorbox{teal!50}{assembly} should form in April to rewrite the constitution, formally end the monarchy and put together the details of the new republican system. The \colorbox{red!50}{\sout{maoists}}
    \colorbox{teal!50}{Maoists} rebels declared a cease fire and signed a peace treaty, agreeing to place its troops and weapons under United Nations supervision.
    & The insurgents would be allowed to return to the government. The treaty establishes that a constitutional \colorbox{red!50}{\sout{asembly}}\colorbox{teal!50}{assembly} should be formed in April to rewrite the constitution, formally end the monarchy and put together the details of the new republican system. The Maoist rebels declared a cease fire and signed a peace treaty, agreeing to place its troops and weapons under U.N. supervision.
    & The insurgents would be allowed to return to the government. \colorbox{red!50}{\sout{It establishes that a}}\colorbox{teal!50}{A} constitutional asembly should form in April to rewrite the constitution, formally end the monarchy and put together the details of the new republican system. \colorbox{red!50}{\sout{Rebels}}\colorbox{teal!50}{The rebels} declared a cease fire and signed a peace treaty, agreeing to place its troops and weapons under U.N. supervision. \\
    \midrule
    3 
    & - 
    & -
    & The insurgents would be allowed to return to the government. \colorbox{red!50}{\sout{A constitutional asembly}}
    \colorbox{teal!50}{Constitutional Assembly} should form in April to rewrite the constitution, formally end the monarchy and put together the details of the new republican system. \colorbox{red!50}{\sout{The rebels}}\colorbox{teal!50}{Rebels} declared a cease fire and signed a peace treaty, agreeing to place its troops and weapons under U.N. supervision. \\
    \bottomrule
  \end{tabular}
  \caption{\label{tab:revision_example}
  A sample snippet of iterative text revisions generated by human writer, \iteraterbase and \method (ours), where $t$ is the revision depth and $t=0$ indicates the original input text.
  Note that \colorbox{red!50}{\sout{text}} represents deletions, and \colorbox{teal!50}{text} represents insertions.
  }
\end{table*}

\end{document}